\documentclass{OAGM}
\OAGMarXiv{1304.1876}
\usepackage{amsmath}
\usepackage{amssymb}
\usepackage{times}
\usepackage{bm}
\usepackage{graphicx}
\newcommand{\subfiglabelright}[1]{%
   \hbox to0pt{\small\kern-2em\hfill\raisebox{.5em}{%
   \colorbox{white}{\hbox to0.7em{\rule{0pt}{0.7em}%
   \hfill\smash{\textbf{#1}}\hfill}}\kern.2em}}}
\title{Algorithmic Optimisations for Iterative Deconvolution Methods}
\author{Martin Welk, Martin Erler\\
University for Health Sciences,
Medical Informatics and Technology (UMIT),\\
Eduard-Walln\"ofer-Zentrum 1, 6060 Hall/Tyrol, Austria}
\begin{document}
\maketitle
\begin{abstract}
We investigate possibilities to speed up iterative algorithms for
non-blind image deconvolution. We focus on algorithms in which 
convolution with the point-spread function to be deconvolved is
used in each iteration, and aim at accelerating these convolution
operations as they are typically the most expensive part of the
computation. We follow two approaches: First, for some practically
important specific point-spread functions, algorithmically efficient
sliding window or list processing techniques can be used. In some
constellations this allows faster computation than via the Fourier 
domain. Second, as iterations progress, computation of convolutions 
can be restricted to subsets of pixels. For moderate thinning rates 
this can be done with almost no impact on the reconstruction quality.
Both approaches are demonstrated in the context of Richardson-Lucy
deconvolution but are not restricted to this method.
\end{abstract}
\section{Introduction}
Sharpening of blurred images by deconvolution has been a long-standing
object of research efforts, and a variety of algorithms for this ill-posed
inverse problem is available which differ widely in their generality, 
restoration quality and computational expense, with articulate trade-offs
between restoration quality and computational cost. 
In the case of space-invariant blur, where each point of
the unobservable (latent) sharp image $g$ is smeared out in the image
plane in the same way, a typical blur model reads as
\begin{equation}
f(x,y) = (g * h)(x,y) + n(x,y) \label{blurmodel}
\end{equation}
where the function $f:\mathbb{R}^2\to\mathbb{R}$ denotes the observed unsharp 
image, $h$ a point-spread function
(PSF) that acts here via convolution $*$, 
and $n$ some additive noise. Dependent on the imaging process, 
the latter may be replaced by other types of noise such as Poisson
or impulse noise.

A so-called \emph{non-blind} deconvolution problem in which the blurred image 
$f$ and the point-spread function $h$ are available as inputs to an algorithm
consists then in determining an image $u$ such that
\begin{equation}
f(x,y) \approx (u*h)(x,y) \;.
\end{equation}

Methods to solve this task range from the fast and simple Wiener filter
\cite{Wiener-Book49} which is essentially a regularised inversion of the
convolution via the Fourier domain, up to expensive iterative methods
\cite{Bar-scs05,Dey-isbi04,Krishnan-nips09,Lucy-AJ74,%
Richardson-JOSA72,Vogel-TIP98,Wang-SIIMS08,Welk-tr10}. 
Some of these algorithms require computation via the Fourier domain,
while others can be implemented in the spatial domain. Only the latter ones
can be extended straightforwardly to the more general setting of spatially
variant blur. 

Except for the Wiener filter, all commonly used algorithms
are iterative. The computational cost of these algorithms is dominated by
convolution operations or Fourier transforms that need to be performed in
each iteration step, and the total number of iterations needed to achieve
satisfactory restoration quality. For many, though not all,
of the iterative methods, the expensive step in
each iteration is a convolution of the current approximation $u^k$ with the 
PSF $h$. This applies, e.g., to total variation deconvolution 
\cite{Chan-TIP98,Vogel-TIP98}
and similar variational methods \cite{Bar-scs05,You-icip96},
Richardson-Lucy deconvolution \cite{Lucy-AJ74,Richardson-JOSA72} and methods 
derived thereof \cite{Dey-isbi04,Welk-tr10}, but not, e.g., to the methods 
from \cite{Krishnan-nips09,Wang-SIIMS08}.

\subsubsection*{Goal and contributions.} We aim at raising the efficiency of
iterative deconvolution algorithms in which convolution with $h$ is carried 
out in each iteration step. To this end, we consider algorithmic improvements
that speed up the computation of convolutions in the spatial domain, without
the use of Fourier transforms. This is interesting for two reasons: First, 
dramatic speedups can be achieved by convolution implementations tailored to 
specific PSFs. This has been demonstrated in a recent paper \cite{Welk-tr12} 
for the particularly favourable case of linear motion blur aligned with an 
axis direction of the sampling grid, in which a fast box-filter algorithm 
clearly outperforms Fourier convolutions. Second, unlike Fourier-based 
algorithms, spatial domain convolution can be generalised to spatially variant
blur settings.

As our first approach, we investigate efficient algorithms for spatial 
convolution with specific types of space-invariant PSFs 
that often occur in practical situations. Thereby, we generalize the
approach of \cite{Welk-tr12}. As our second approach, we address the 
possibility to
restrict convolution to subsets of image pixels in order
to save computation cost on those pixels which do not change anymore.

By describing our work as \emph{algorithmic optimisations}, we follow
the widespread use of this term in computer science for algorithmic
modifications that allow to perform a computation in a more resource-efficient
way (although no optimum in the mathematical sense is usually achieved), 
see e.g.\ \cite{Gerdt-PCS02} or for the slightly more general term
\emph{program optimisation} (which also includes code optimisation)
\cite[p.~84]{Sedgewick-Book84}.

\subsubsection*{Related work.} Efficient box filtering goes back to 
McDonnell \cite{McDonnell-CGIP81}, but seems to have received little
attention in the image deconvolution context.
Efficient convolution techniques, especially for
kernels with uniform intensity, have also been investigated in
\cite{Mount-TIP01}. For efforts to accelerate RL deconvolution see
e.g.\ \cite{Biggs-AO97,Holmes-JOSAA91}.

\subsubsection*{Structure of the Paper.} We recall Richardson-Lucy deconvolution
as the underlying iterative method in Section~\ref{sec-rl}. The first
approach to algorithmic improvements, addressing specifically structured
convolution kernels, is the topic
of Section~\ref{sec-fastconv}. The second approach, restricting
convolution computation to subsets, is considered in Section~\ref{sec-thin}.
The achievable speed-ups are demonstrated by experiments in 
Section~\ref{sec-exp}, followed by conclusions in Section~\ref{sec-conc}.

\section{Richardson-Lucy Deconvolution}\label{sec-rl}

Richardson-Lucy deconvolution (RL) \cite{Lucy-AJ74,Richardson-JOSA72}
is widely used because of its simplicity and favourable results for moderate
noise. Starting from $u^0:=f$, it uses the iteration
\begin{equation}
u_{k+1} = \left(h^* * \frac{f}{u^k*h}\right)\cdot u^k
\end{equation}
to create a sequence
$(u^0,u^1,u^2,\ldots)$ of images of increasing sharpness.
Here, $h^*$ denotes the
adjoint PSF, geometrically obtained by 
reflecting $h$ about the origin. The number of iterations until
stopping is the single parameter of the method.

Although we present only tests with RL here, our algorithmic improvements
can be applied equally to modified RL methods with additional regularisers
or robustified data terms \cite{Dey-isbi04,Welk-tr10}, with similar
results (compare \cite{Welk-tr12} for linear motion blur).

\subsubsection*{Boundary treatment.} Convolution with (spatially invariant)
PSF always transfers information across image boundaries. In turn,
missing information in deconvolution typically leads to artifacts 
propagating from the boundary into the interior of the image.
There are two basic ways to handle these boundary problems in computing
convolutions. Either, convolution is computed on a reduced domain for 
which all input data are available. This, however, is not an option in 
iterative procedures. Alternatively, the image has to be continued (padded). 
Convolution via discrete Fourier transforms implicitly
introduces periodic continuation. In spatial domain convolution, an often 
reasonable compromise between suppression of
artifacts and computational effort is 
to replace every missing input pixel
(outside the image bounds) with the closest image pixel, which means 
a constant continuation along lines perpendicular to the image boundary.
We will implement this boundary treatment in all of our convolution
procedures acting in the spatial domain.

\section{Fast Convolution with Special Point-Spread Functions}
\label{sec-fastconv}

\begin{figure}[t!]
\centering\begin{tabular}{@{}c@{~~~}c@{~~~}c@{~~~}c@{~~~}c@{}}
(a) \includegraphics[width=0.12\textwidth]{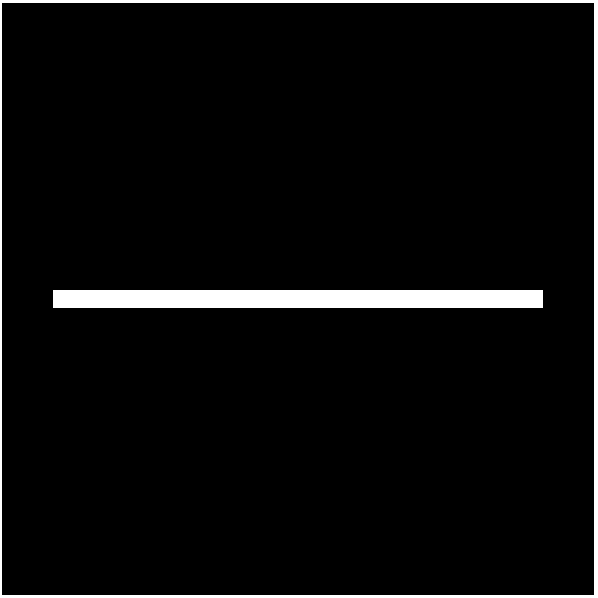}&
(b) \includegraphics[width=0.12\textwidth]{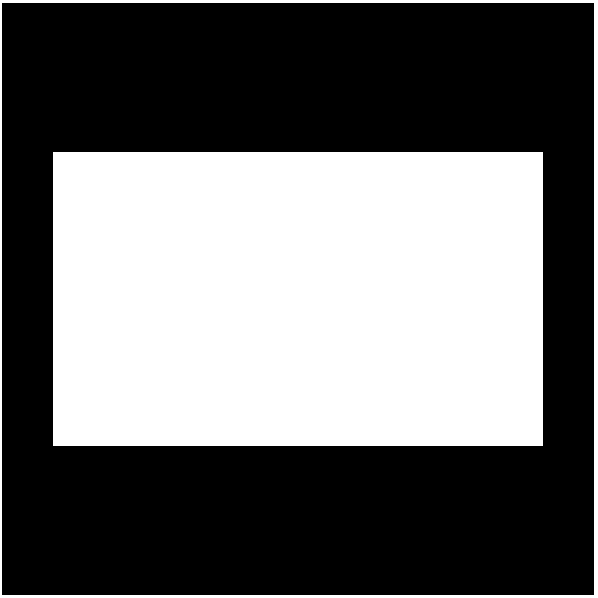}&
(c) \includegraphics[width=0.12\textwidth]{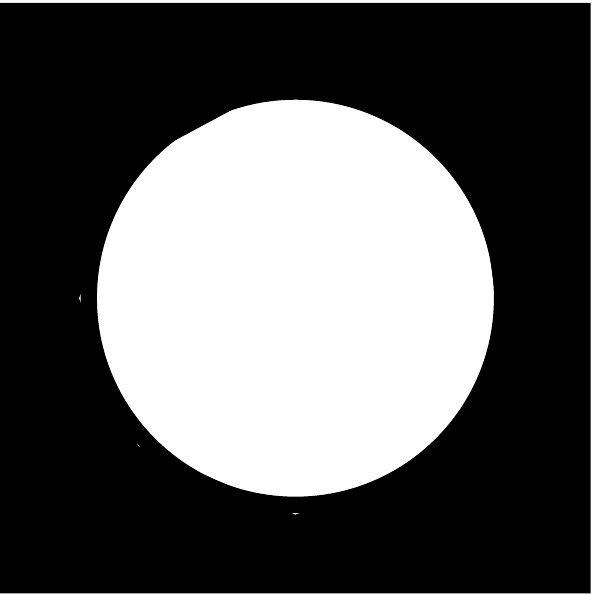}&
(d) \includegraphics[width=0.12\textwidth]{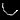}&
(e) \includegraphics[width=0.12\textwidth]{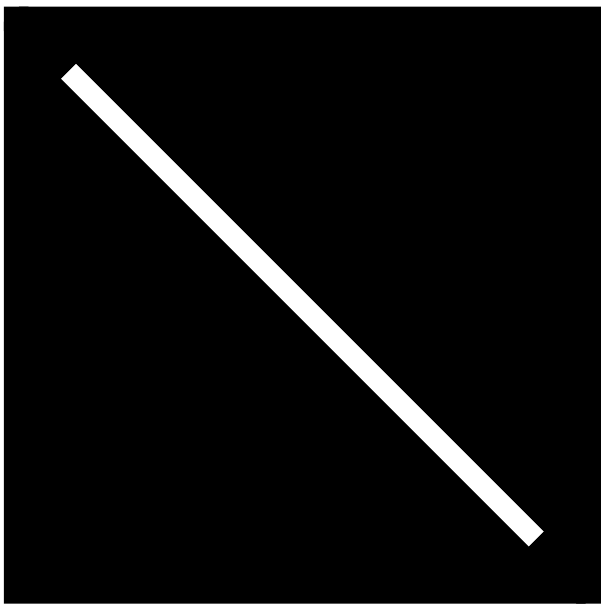}
\end{tabular}\\
\caption{Point spread functions. \textbf{Left to right:} 
\textbf{(a)} Linear motion blur along scan-lines. -- 
\textbf{(b)} Rectangular PSF. -- 
\textbf{(c)} Defocus blur. -- 
\textbf{(d)} Sparse PSF similar to camera shake. -- 
\textbf{(e)} Diagonal line blur.}
\label{fig-psfs}
\end{figure}

\subsubsection*{Linear motion blur.}
The simplest case 
we mention is
linear motion blur in scan-line ($x$) direction, see
Figure~\ref{fig-psfs}(a). 
This particular PSF has already been considered in \cite{Welk-tr12}. Here,
convolution can efficiently be implemented using a fast box filter 
\cite{McDonnell-CGIP81} that acts in each scan-line via a sliding window. 
For an $N_x\times N_y$-image and an $M$-pixel PSF, the complexity of this 
filter is $\mathcal{O}(N_y\cdot(N_x+M))$, as contrasted to
$\mathcal{O}(N_xN_yM)$ for naive spatial convolution or 
$\mathcal{O}(N_xN_y\log(N_xN_y))$ for an FFT-based method.

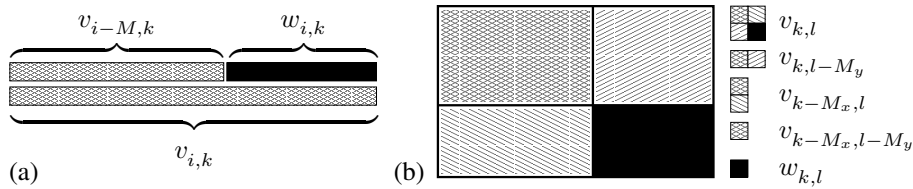
\begin{figure}[t!]
\centering\begin{tabular}{@{}c@{~~}c@{}}
\unitlength0.001\textwidth
\begin{picture}(400,190)
\put(0,0){(a)}
\put(0,80){\includegraphics[width=0.40\textwidth]{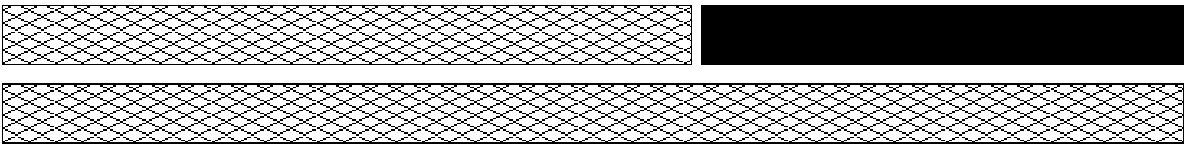}}
\put(2,73){$\underbrace{\rule{0.398\textwidth}{0pt}}
_{\textstyle v^{\phantom{M}}_{i,k}}$}
\put(2,130){$\overbrace{\rule{0.23\textwidth}{0pt}}
^{\textstyle v^{\phantom{M}}_{i-M,k}}$}
\put(238,130){$\overbrace{\rule{0.16\textwidth}{0pt}}
^{\textstyle w^{\phantom{M}}_{i,k}}$}
\end{picture}&
\unitlength0.001\textwidth
\begin{picture}(550,190)
\put(0,0){(b)}
\put(45,0){\includegraphics[width=0.36\textwidth]{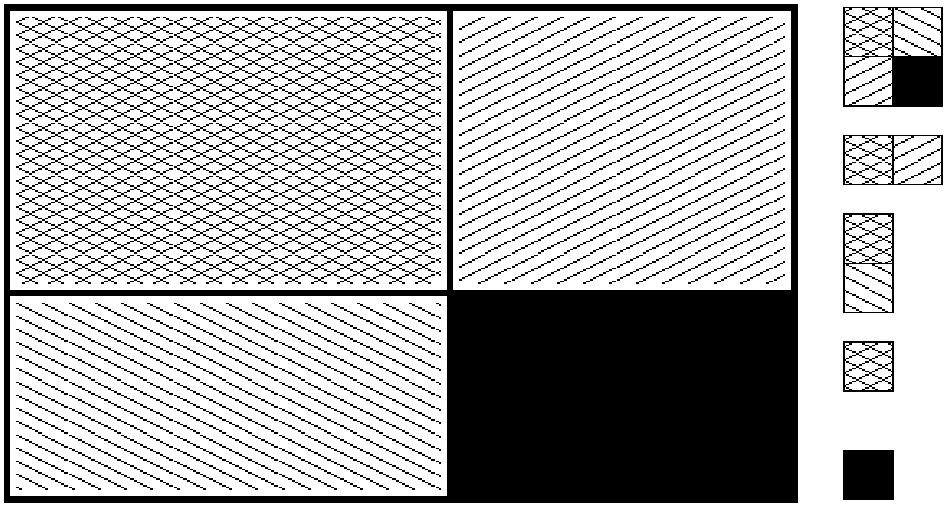}}
\put(415,165){$v^{\phantom{M}}_{k,l}$}
\put(415,125){$v^{\phantom{M}}_{k,l-M_y}$}
\put(415,85){$v^{\phantom{M}}_{k-M_x,l}$}
\put(415,45){$v^{\phantom{M}}_{k-M_x,l-M_y}$}
\put(415,5){$w^{\phantom{M}}_{k,l}$}
\end{picture}
\end{tabular}\\
\caption{Box filter computation via cumulated sums, 
\textbf{(a)} in 1D, \textbf{(b)} in 2D.}
\label{fig-boxcum}
\end{figure}

The same result, with basically the same complexity, can be obtained as 
follows: one computes first in each scan-line $u_{*,k}$ of a discrete image
$u=(u_{i,j})$ an array of 
cumulated sums $v_{i,k}:=\sum_{j=1}^iu_{i,k}$; then each pixel of the 
convolution result $w=u*h$ is given by subtracting two array entries such 
as $w_{i,k}=v_{i,k}-v_{i-M,k}$, compare Fig.~\ref{fig-boxcum}(a). 
Each of the two steps has linear complexity in the number of pixels; 
however, to implement the desired boundary conditions, the image 
must be extended by the PSF size in $x$ direction, implying again 
$\mathcal{O}(N_y\cdot(N_x+M))$ overall complexity.

\subsubsection*{Rectangular blur.}
As our next test case we consider a 2-dimensional PSF consisting of a 
rectangle with constant density aligned with the scan-lines, see 
Fig.~\ref{fig-psfs}(b).
While this PSF type is of less practical importance, it can serve as an 
intermediate step towards the more realistic PSFs considered later on.
A straightforward transfer of the sliding window idea uses the separability
of this PSF: an intermediate image is computed by the box-filter method
in $x$ direction, followed by another box-filter step in $y$ direction.
The overall complexity of this approach for an $N_x\times N_y$-image and
an $M_x\times M_y$-pixel PSF is 
$\mathcal{O}(N_y\cdot(N_x+M_x)+(N_x+M_x)\cdot(N_y+M_y))
=\mathcal{O}((N_x+M_x)(N_y+M_y))$.

An alternative generalisation starts from the above-mentioned cumulated
sum approach. In a first step, a cumulated sum array $(v_{k,l})$ is computed,
in which $v_{k,l}=\sum_{i=1}^k\sum_{j=1}^lu_{i,j}$ contains the sum
of all grey-values left and above position $(k,l)$. This step has linear
complexity in the number of pixels. In a second step, one computes
each pixel of $w=u*h$ via
$w_{k,l}=v_{k,l}-v_{k-M_x,l}-v_{k,l-M_y}+v_{k-M_x,l-M_y}$, compare
Fig.~\ref{fig-boxcum}(b).
Again, for the desired boundary conditions, the cumulated image must be
computed in size $(N_x+M_x)(N_y+M_y)$, yielding the same overall complexity
$\mathcal{O}((N_x+M_x)(N_y+M_y))$ as above. 
One checks easily, however, that
less operations per pixel are needed.

\subsubsection*{Defocus blur.} 
A frequent source of blur in image acquisition is defocussing which, for a 
circular-shaped camera aperture, leads to a disc-shaped point spread
with constant density. 
To apply the sliding window approach, 
note that shifting by one pixel to the right removes from the mask 
just one (no longer straight) line of pixels at the left boundary
while adding one line at the right boundary. If the 
PSF is $M_y$ pixels high,
this update requires $2 M_y$ operations.
For a PSF enclosed in an $M_x\times M_y$ box, the complete
sliding window summation thus takes $\mathcal{O}(N_yM_y(N_x+M_x))$ 
time. This algorithm can be applied to any convex shape with uniform
intensity. We will refer to it as \emph{generic box filter.} 
The cumulated sum approach does not reduce complexity in this
situation.

\subsubsection*{Sparse blur.}
If images are degraded by motion blur not aligned to the sensor grid, or 
irregular motion e.g.\ due to
camera shake, representing the point-spread function in an axis-parallel 
rectangle means that most pixels will be zero. Direct summation over
such an enclosing rectangle is therefore inefficient.

An alternative representation of such a PSF is a list of
tuples where each tuple contains coordinates and intensity of one support
pixel of the PSF. By summation over this list, convolution is computed
in $\mathcal{O}(N_xN_yM)$ time where $M$ is the number of support
pixels. This procedure will be denoted as \emph{list filter.}
Unlike in the previous settings, we do not specifically consider
PSFs with uniform intensity on all support pixels, since sparse blurs
rarely satisfy such a condition.

\section{Selective Convolution}\label{sec-thin}

In RL deconvolution each single iteration acts locally: It measures the 
error of the re-blurred image $u*h$ compared to the observed image $f$ 
(by the quotient $f/(u*h)$), and redistributes this error by
another blurring step (with the adjoint PSF $h^*$). Other iterative
deconvolution methods that use convolution with the given PSF work in
a similar way. Such a process can take many iterations
to finally transport the error corrections to their correct locations, and
achieve a good overall reconstruction.

Observation shows, however, that often the sharpness of the iterates
improves dramatically during the first few iteration steps, and large
parts of the image (typically, those with simpler structures) are well 
reconstructed already at this point. It is only a minority of the pixels
in more complex structured image regions that cause the demand for
many iterations. Based on this observation, it appears attractive to
focus the computation to those pixels which really need the work, and
to spare those pixels which are not or almost not changed during later
iterations.

We implemented this idea in a very simple form:
In each iteration, the absolute changes of all pixel values are checked,
and those pixels whose change is below a given threshold are marked
\emph{inactive} in subsequent iterations, and thereby excluded from the
expensive convolutions. For RL, this comes down to
\begin{align}
v_{i,j}^k &:= (u^k*h)_{i,j}\;,&u_{i,j}^{k+1} &:= (h^* * (f/v^k))_{i,j}\cdot u_{i,j}^k\;,&
&\text{if $(i,j)$ is active,}\\
v_{i,j}^k &:= v_{i,j}^{k-1}\;,&u_{i,j}^{k+1} &:= u_{i,j}^k\;,&
&\text{if $(i,j)$ is inactive.}
\end{align}
Every ten iterations, all pixels are set back to \emph{active}, thus one full 
iteration step is carried out, allowing to bring
pixels back into the update process which still need small updates under
the influence of more active pixels in their neighbourhood.

While more complicated adaptive update rules can be conceived, and are
worth further research, an advantage of the present very simple
rule is that the selection process itself requires almost no 
computational effort, which will also be demonstrated
experimentally in the next section.

Selective convolution cannot be combined with those fast convolution
techniques from Section~\ref{sec-fastconv} that rely on sliding windows or
cumulated sums. However, it can be combined with the list filter.

\section{Experimental Evaluation}\label{sec-exp}

All algorithms were implemented in C, and compiled with gcc 4.6.3 at 
optimisation level O2.
Computations were run single-threaded on a PC with an 
AMD Phenom X4 quad-core 64-bit CPU clocked at 3.00 GHz under Ubuntu Linux 
12.04. 
As this is not a real-time environment, interference of other processes
in the system causes stochastic variations in runtime. Therefore
each runtime was averaged from a series of 100 program runs,
which reduced the standard deviation to 
$0.01$ seconds or less in each series.

\subsubsection*{Specific convolution operators.}
In our first experiment, Table~\ref{tab-rt1}, we measure the computation
time of 100 RL iterations with the different convolution implementations
from Section~\ref{sec-fastconv}. For reference, naive spatial convolution
(direct summation over an axis-parallel rectangle enclosing the PSF 
support) and an implementation via Fast Fourier Transform
are included. The moderate PSF dimensions of $9$ and $17$ pixels
correspond to common practical use cases.

\begin{table}[b!]
\centering
\caption{Runtime comparison of RL deconvolution with different
PSFs and different convolution algorithms. 
Each value is the average runtime in seconds from 100 program runs,
with 100 iterations each, on a 3 GHz CPU.
Further details see text.}
\label{tab-rt1}
\medskip

\begin{tabular}{|l|c@{~~}c|c@{~~}c|c@{~~}c|c@{~~}c|}
\hline\rule{0pt}{1.3em}\raisebox{0.2em}{PSF type}
&\multicolumn{2}{c|}{\smash{\raisebox{-0.3em}{%
\includegraphics[width=1.4em]{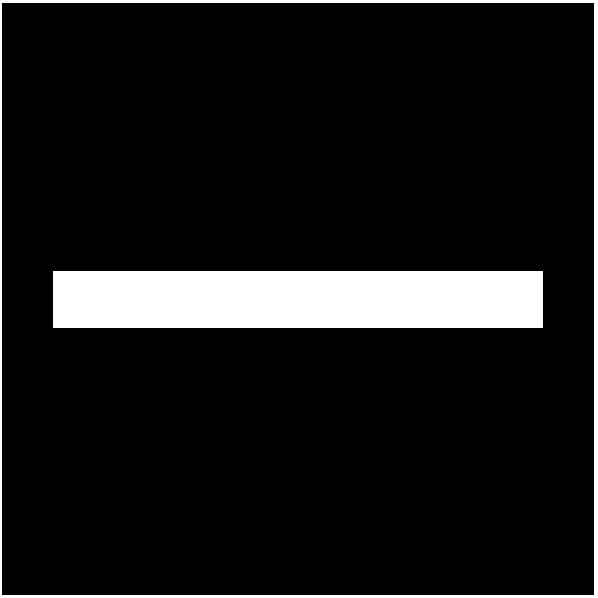}}}}
&\multicolumn{2}{c|}{\smash{\raisebox{-0.3em}{%
\includegraphics[width=1.4em]{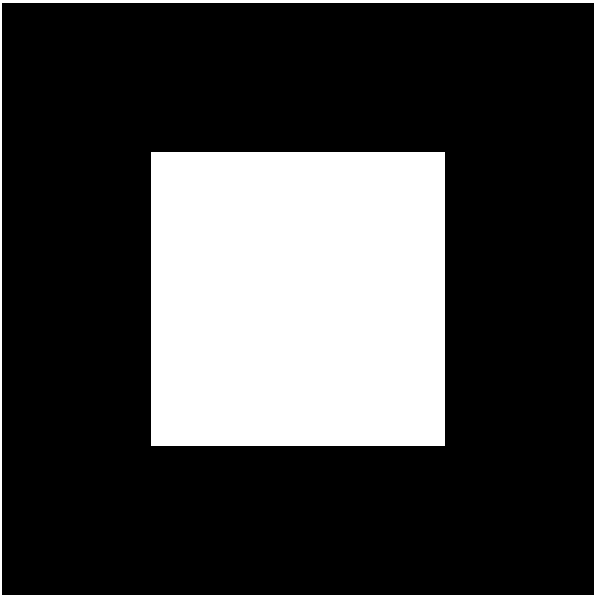}}}}
&\multicolumn{2}{c|}{\smash{\raisebox{-0.3em}{%
\includegraphics[width=1.4em]{images/fig1e.pdf}}}}
&\multicolumn{2}{c|}{\smash{\raisebox{-0.3em}{%
\includegraphics[width=1.4em]{images/fig1c.pdf}}}}\\
PSF size
&$9$&$17$&$9^2$&$17^2$&$9$&$17$&
$\varnothing~ 9$&$\varnothing~ 17$\\
\hline
Naive spatial&0.48&0.72&3.74&10.90&3.74&10.90&3.74&10.90\\
Fourier &1.74&1.75&1.75&1.75&1.75&1.74&1.74&1.75\\
List&0.50&0.78&3.14&10.73&\textbf{0.51}&\textbf{0.80}&1.97&7.37\\
Generic box&0.17&0.17&0.93&1.56&0.91&1.48&\textbf{0.93}&\textbf{1.54}\\
Box 2D sliding&0.13&0.13&0.70&1.19&---&---&---&---\\
Box 2D cumul.&0.21&0.22&\textbf{0.21}&\textbf{0.23}&---&---&---&---\\
Box 1D&\textbf{0.09}&\textbf{0.09}&---&---&---&---&---&---\\
\hline
\end{tabular}
\end{table}

\begin{figure}[t!]
\centering
\begin{tabular}{@{}c@{~}c@{~}c@{}}
\includegraphics[width=0.32\textwidth]{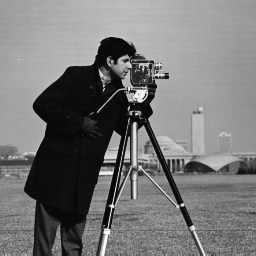}%
\subfiglabelright{a}&
\includegraphics[width=0.32\textwidth]{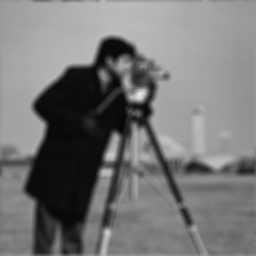}%
\subfiglabelright{b}&
\includegraphics[width=0.32\textwidth]{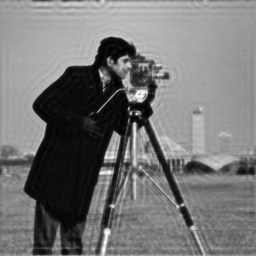}%
\subfiglabelright{c}\\
&
\includegraphics[width=0.32\textwidth]{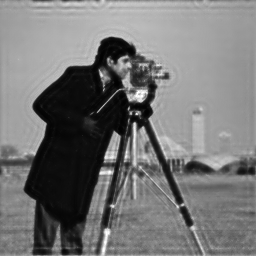}%
\subfiglabelright{d}&
\includegraphics[width=0.32\textwidth]{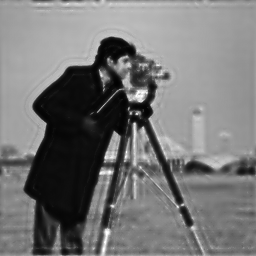}%
\subfiglabelright{e}
\end{tabular}
\caption{Richardson-Lucy deconvolution with selective convolution.
\textbf{Top left to bottom right in rows:}
\textbf{(a)} Original image, $256\times256$ pixels. --
\textbf{(b)} Blurred with defocus PSF, diameter $9$ pixels. --
\textbf{(c)} Reference result from 100 iterations of standard RL
(SNR: 13.31 dB). --
\textbf{(d)} Same but with thinning threshold $0.1$ (SNR: 13.11 dB). --
\textbf{(e)} With thinning threshold $0.5$ (SNR: 12.23 dB).}
\label{fig-sparse}
\end{figure}

For each experiment, a test image was generated by convolving the 
$256\times256$ \emph{cameraman} image, see Fig.~\ref{fig-sparse}(a),
with one of the following eight point spread
functions: 1D box representing a linear motion blur in $x$ direction, 9 or
17 pixels long; 2D square-shaped box kernel of $9\times9$ or $17\times17$
pixels; diagonal ($45^\circ$) line kernel with $9$ or $17$ pixels, as a simple 
representative of a sparse point spread function; defocus kernel
of $9$ or $17$ pixels diameter. Defocus blurring with diameter $9$ is
shown in Fig.~\ref{fig-sparse}(b). As the convolution was carried out via the
Fourier domain, periodic boundary conditions led to slight wrap-around
artifacts.
Each test image was RL-deconvolved with its matching PSF, testing all those
convolution routines which could handle the respective PSF.

As expected, in all cases the most specific applicable convolution algorithms 
lead to the fastest deconvolution. The generic box algorithm 
performs favourable for all ``massive'' convex 2D PSFs of constant density, 
while sparse PSFs are well treated by the list filter. Both implementations
outperform FFT-based methods in their respective application
areas. However, for the 2D box and defocus PSF of edge length/diameter $17$ 
the FFT method is close to break even with the generic box implementation.

\begin{table}[b!]
\centering
\caption{Runtime and reconstruction quality for
thinned RL deconvolution of the cameraman image blurred with 
$\varnothing~9$ defocus PSF.
Runtimes are averaged from 100 program runs,
with 100 iterations each, 
on a 3 GHz CPU.
Further details see text.}
\label{tab-rt2}
\medskip

\begin{tabular}{|l|c|c|c|c|c|}
\hline
Threshold&Omitted&\multicolumn{2}{c|}{SNR (dB)}&
time&speedup\\
&pixels (\%)&vs. orig.&vs. ref.&(s)&\\
\hline
Reference& \phantom00.00&13.31&  ---&3.74&1.00\\
\hline
$\quad0$    &\phantom00.00&13.31&  ---&3.77&0.99\\
$\quad0.005$&12.54&13.30&45.33&3.34&1.12\\
$\quad0.01$ &22.32&13.31&42.32&2.99&1.25\\
$\quad0.02$ &34.58&13.30&39.54&2.55&1.47\\
$\quad0.05$ &53.69&13.25&34.74&1.84&2.03\\
$\quad0.1$  &66.34&13.11&30.82&1.36&2.75\\
$\quad0.2$  &75.57&12.84&27.18&1.01&3.70\\
$\quad0.5$  &83.48&12.23&22.67&0.71&5.27\\
\hline
\end{tabular}
\end{table}

\subsubsection*{Selective convolution.}
In our second experiment, Table~\ref{tab-rt2} and Fig.~\ref{fig-sparse},
we performed RL deconvolution with the naive spatial convolution but
the convolution was carried out only for part of the pixels, as described
in Section~\ref{sec-thin}. The cameraman image convolved with the defocus
PSF of diameter $9$ served as test case. For this test case 100 iterations
of RL lead to a visible sharpening.

In Table~\ref{tab-rt2}, we report for different values of the thinning
threshold the ratios of omitted pixels in convolutions, two signal-to-noise
ratios, run times (again averaged from 100 runs each) and speedup factors.
The SNR value w.r.t.\ the original cameraman image allows to assess
the reconstruction quality for each threshold, while SNR w.r.t.\ the
reference image (100 iterations of unthinned RL) measures deviations
introduced by the thinning.

The first line of the table contains the reference values for
the RL method without thinning. Using the implementation with thinning
but with the threshold set to zero (next line) reveals that the cost of
the additional logic for thinning is not more than about $1\,\%$. Raising
the threshold up to $0.1$ speeds up the computation up to about 2.75 times
while the reconstruction quality is almost not affected, as can be seen from
the SNR figures, and is visually confirmed in Fig.~\ref{fig-sparse}(d).
With larger thresholds and thus more aggressive thinning, deconvolution 
results are visibly affected, see Fig.~\ref{fig-sparse}(e).

\section{Conclusions}\label{sec-conc}

We have demonstrated that the computational expense of iterative
deconvolution methods can be substantially reduced if the point spread
function is of one of several specific types that occur often in practice.
The way to achieve this is to use specialised convolution operators.
For small blurs, computation can be several times as fast as with 
Fourier-based convolution. Moreover, we have seen that some speed-up
(in our example about 2.5) can also be achieved by selectively performing
convolution operations in later iteration steps only for image pixels where 
still significant change takes place.

From the deconvolution point of view, our experimental setting is prone
to over-assess reconstruction quality. In the context of the present paper,
which is not at all about reconstruction quality but only concerned with
algorithmic optimisations, this is not an issue, however.

A deeper investigation of the selective convolution approach including
refined thinning rules, a broader experimental evaluation, and combination 
with the list filter is the subject of
ongoing work. GPU-based parallelisation of the specialised convolution 
filters is another interesting topic for further research, as well as the
integration of the techniques in blind or semi-blind deconvolution 
frameworks.

\end{document}